\documentclass{sig-alternate-05-2015}
\usepackage{booktabs}

\begin{document}
\makeatletter
\def\@copyrightspace{\relax}
\makeatother

\title{Wide \& Deep Learning for Recommender Systems}

\numberofauthors{1}

\author{
\alignauthor
Heng-Tze Cheng,
Levent Koc,
Jeremiah Harmsen,
Tal Shaked,
Tushar Chandra,
Hrishi Aradhye,
Glen Anderson,
Greg Corrado,
Wei Chai,
Mustafa Ispir,
Rohan Anil,
Zakaria Haque,
Lichan Hong,
Vihan Jain,
Xiaobing Liu,
Hemal Shah\\
\affaddr{Google Inc.\titlenote{Corresponding author: \texttt{hengtze@google.com}}}
}

\maketitle
\begin{abstract}
Generalized linear models with nonlinear feature transformations are widely used for large-scale regression and classification problems with sparse inputs. Memorization of feature interactions through a \textit{wide} set of cross-product feature transformations are effective and interpretable, while generalization requires more feature engineering effort. With less feature engineering, \textit{deep} neural networks can generalize better to unseen feature combinations through low-dimensional dense embeddings learned for the sparse features. However, deep neural networks with embeddings can over-generalize and recommend less relevant items when the user-item interactions are sparse and high-rank. In this paper, we present Wide \& Deep learning---jointly trained wide linear models and deep neural networks---to combine the benefits of memorization and generalization for recommender systems. We productionized and evaluated the system on Google Play, a commercial mobile app store with over one billion active users and over one million apps. Online experiment results show that Wide \& Deep significantly increased app acquisitions compared with wide-only and deep-only models. We have also open-sourced our implementation in TensorFlow.
\end{abstract}




\section{Introduction}
A recommender system can be viewed as a search ranking system, where the input query is a set of user and contextual information, and the output is a ranked list of items. Given a query, the recommendation task is to find the relevant items in a database and then rank the items based on certain objectives, such as clicks or purchases.

One challenge in recommender systems, similar to the general search ranking problem, is to achieve both \textit{memorization} and \textit{generalization}.
Memorization can be loosely defined as learning the frequent co-occurrence of items or features and exploiting the correlation available in the historical data. Generalization, on the other hand, is based on transitivity of correlation and explores new feature combinations that have never or rarely occurred in the past.
Recommendations based on memorization are usually more topical and directly relevant to the items on which users have already performed actions.
Compared with memorization, generalization tends to improve the diversity of the recommended items. In this paper, we focus on the apps recommendation problem for the Google Play store, but the approach should apply to generic recommender systems.

For massive-scale online recommendation and ranking systems in an industrial setting, generalized linear models such as logistic regression are widely used because they are simple, scalable and interpretable. The models are often trained on binarized sparse features with one-hot encoding. E.g., the binary feature ``\texttt{user\_installed\_app=netflix}'' has value 1 if the user installed Netflix. Memorization can be achieved effectively using cross-product transformations over sparse features, such as  \texttt{AND}(\texttt{user\_installed\_app=netflix}, \texttt{impression\_app=pandora}''), whose value is 1 if the user installed Netflix and then is later shown Pandora. This explains how the co-occurrence of a feature pair correlates with the target label. Generalization can be added by using features that are less granular, such as \texttt{AND}(\texttt{user\_installed\_category=video}, \texttt{impression\_category=music}), but manual feature engineering is often required. One limitation of cross-product transformations is that they do not generalize to query-item feature pairs that have not appeared in the training data.

Embedding-based models, such as factorization machines \cite{LibFMTIST12} or deep neural networks, can generalize to previously unseen query-item feature pairs by learning a low-dimensional dense embedding vector for each query and item feature, with less burden of feature engineering. However, it is difficult to learn effective low-dimensional representations for queries and items when the underlying query-item matrix is sparse and high-rank, such as users with specific preferences or niche items with a narrow appeal. In such cases, there should be no interactions between most query-item pairs, but dense embeddings will lead to nonzero predictions for all query-item pairs, and thus can over-generalize and make less relevant recommendations. On the other hand, linear models with cross-product feature transformations can memorize these ``exception rules'' with much fewer parameters.

\begin{figure*}[t!]
	\centering
	\includegraphics[width=\textwidth, trim = 0 10 0 10]{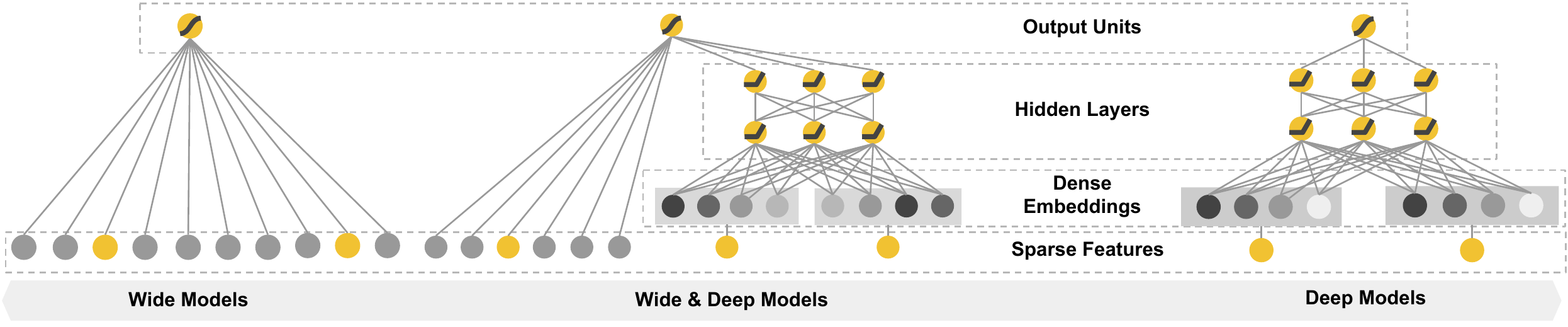}
	\caption{The spectrum of Wide \& Deep models.}
	\label{fig:WideDeepSpectrum}
\end{figure*}

In this paper, we present the Wide \& Deep learning framework to achieve both memorization and generalization in one model, by jointly training a linear model component and a neural network component as shown in Figure \ref{fig:WideDeepSpectrum}.

The main contributions of the paper include:
\begin{itemize}
\item The Wide \& Deep learning framework for jointly training feed-forward neural networks with embeddings and linear model with feature transformations for generic recommender systems with sparse inputs.
\item The implementation and evaluation of the Wide \& Deep recommender system productionized on Google Play, a mobile app store with over one billion active users and over one million apps.
\item We have open-sourced our implementation along with a high-level API in TensorFlow\footnote{See Wide \& Deep Tutorial on \texttt{http://tensorflow.org}.}.
\end{itemize}

While the idea is simple, we show that the Wide \& Deep framework significantly improves the app acquisition rate on the mobile app store, while satisfying the training and serving speed requirements.

\section{Recommender System Overview}

\begin{figure}[t!]
	\centering
	\includegraphics[width=\columnwidth, trim = 0 20 0 20]{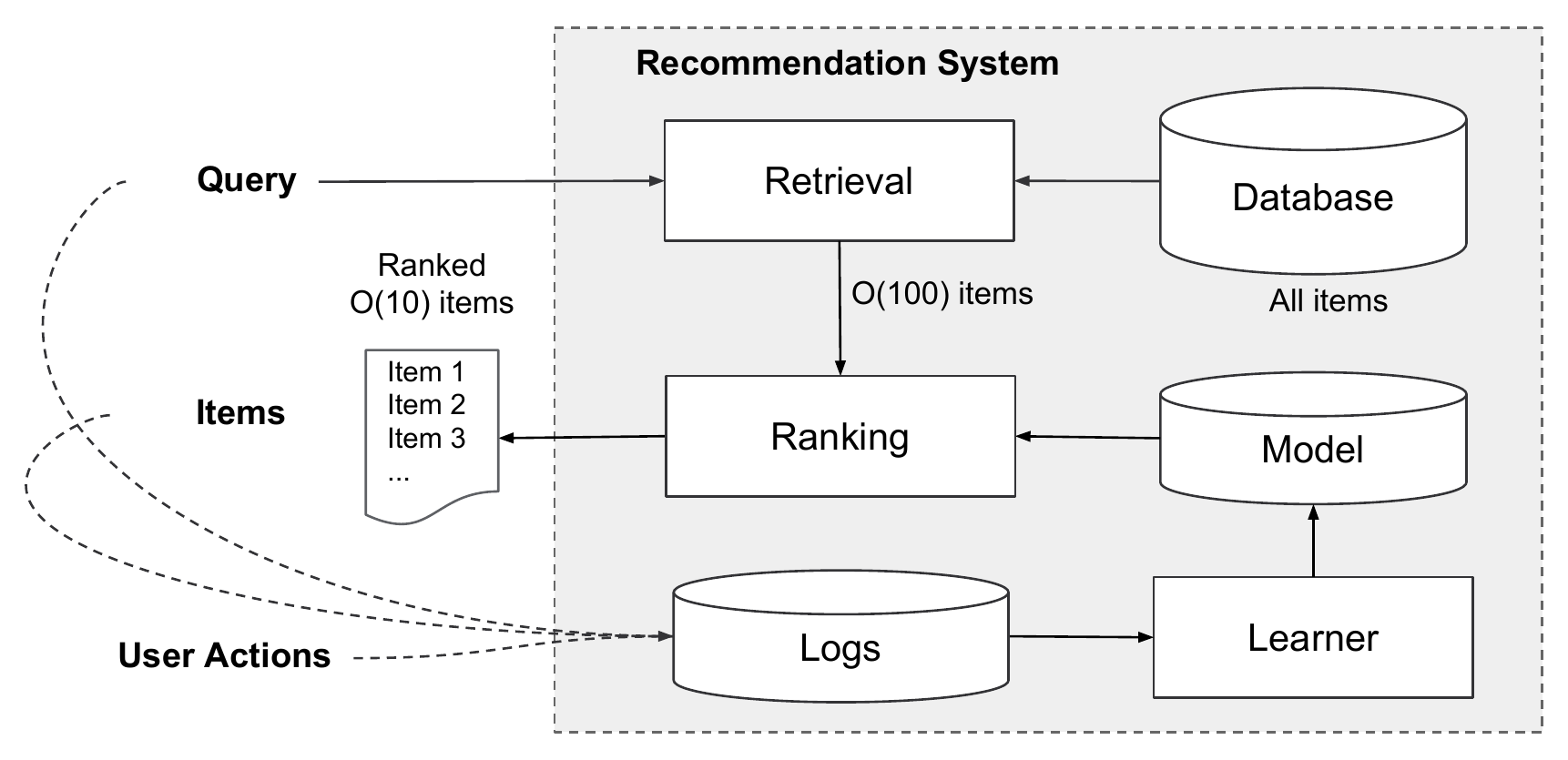}
	\caption{Overview of the recommender system.}
	\label{fig:RecommenderSystemOverview}
\end{figure}

An overview of the app recommender system is shown in Figure \ref{fig:RecommenderSystemOverview}. A query, which can include various user and contextual features, is generated when a user visits the app store. The recommender system returns a list of apps (also referred to as impressions) on which users can perform certain actions such as clicks or purchases. These user actions, along with the queries and impressions, are recorded in the logs as the training data for the learner.

Since there are over a million apps in the database, it is intractable to exhaustively score every app for every query within the serving latency requirements (often $O(10)$ milliseconds). Therefore, the first step upon receiving a query is \textit{retrieval}. The retrieval system returns a short list of items that best match the query using various signals, usually a combination of machine-learned models and human-defined rules. After reducing the candidate pool, the \textit{ranking} system ranks all items by their scores. The scores are usually $P(y|\mathbf{x})$, the probability of a user action label $y$ given the features $\mathbf{x}$, including user features (e.g., country, language, demographics), contextual features (e.g., device, hour of the day, day of the week), and impression features (e.g., app age, historical statistics of an app). In this paper, we focus on the ranking model using the Wide \& Deep learning framework.

\section{Wide \& Deep Learning}

\subsection{The Wide Component}
The wide component is a generalized linear model of the form $y = \mathbf{w}^T\mathbf{x}+b$, as illustrated in Figure \ref{fig:WideDeepSpectrum} (left). $y$ is the prediction, $\mathbf{x}=[x_1, x_2, ..., x_d]$ is a vector of $d$ features, $\mathbf{w}=[w_1, w_2, ..., w_d]$ are the model parameters and $b$ is the bias. The feature set includes raw input features and transformed features. One of the most important transformations is the \textit{cross-product transformation}, which is defined as:
\begin{equation}
\phi_k(\mathbf{x}) = \prod_{i=1}^{d} x_i^{c_{ki}} \quad c_{ki} \in \{0,1\}
\end{equation}
where $c_{ki}$ is a boolean variable that is 1 if the $i$-th feature is part of the $k$-th transformation $\phi_k$, and 0 otherwise. For binary features, a cross-product transformation (e.g., ``\texttt{AND}(\texttt{gender=female}, \texttt{language=en})'') is 1 if and only if the constituent features (``\texttt{gender=female}'' and ``\texttt{language=en}'') are all 1, and 0 otherwise.
This captures the interactions between the binary features, and adds nonlinearity to the generalized linear model.

\subsection{The Deep Component}
The deep component is a feed-forward neural network, as shown in Figure \ref{fig:WideDeepSpectrum} (right).
For categorical features, the original inputs are feature strings (e.g., ``\texttt{language=en}''). Each of these sparse, high-dimensional categorical features are first converted into a low-dimensional and dense real-valued vector, often referred to as an embedding vector. The dimensionality of the embeddings are usually on the order of $O(10)$ to $O(100)$. The embedding vectors are initialized randomly and then the values are trained to minimize the final loss function during model training. These low-dimensional dense embedding vectors are then fed into the hidden layers of a neural network in the forward pass. Specifically, each hidden layer performs the following computation:
\begin{equation}
a^{(l+1)} = f(W^{(l)} a^{(l)} + b^{(l)})
\end{equation}
where $l$ is the layer number and $f$ is the activation function, often rectified linear units (ReLUs). $a^{(l)}$, $b^{(l)}$, and $W^{(l)}$ are the activations, bias, and model weights at $l$-th layer.

\subsection{Joint Training of Wide \& Deep Model}
The wide component and deep component are combined using a weighted sum of their output log odds as the prediction, which is then fed to one common logistic loss function for joint training.
Note that there is a distinction between \textit{joint training} and \textit{ensemble}.
In an ensemble, individual models are trained separately without knowing each other, and their predictions are combined only at inference time but not at training time.
In contrast, joint training optimizes all parameters simultaneously by taking both the wide and deep part as well as the weights of their sum into account at training time.
There are implications on model size too:
For an ensemble, since the training is disjoint, each individual model size usually needs to be larger (e.g., with more features and transformations) to achieve reasonable accuracy for an ensemble to work.
In comparison, for joint training the wide part only needs to complement the weaknesses of the deep part with a small number of cross-product feature transformations, rather than a full-size wide model.

Joint training of a Wide \& Deep Model is done by back-propagating the gradients from the output to both the wide and deep part of the model simultaneously using mini-batch stochastic optimization. In the experiments, we used Follow-the-regularized-leader (FTRL) algorithm \cite{FTRL11} with $L_1$ regularization as the optimizer for the wide part of the model, and AdaGrad \cite{Adagrad11} for the deep part.

The combined model is illustrated in Figure \ref{fig:WideDeepSpectrum} (center).
For a logistic regression problem, the model's prediction is:
\begin{equation}
P(Y=1|\mathbf{x}) = \sigma(\mathbf{w}_{wide}^T[\mathbf{x}, \mathbf{\phi(x)}] + \mathbf{w}_{deep}^T a^{(l_f)} + b)
\end{equation}
where $Y$ is the binary class label, $\sigma(\cdot)$ is the sigmoid function, $\mathbf{\phi(x)}$ are the cross product transformations of the original features $\mathbf{x}$, and $b$ is the bias term. $\mathbf{w}_{wide}$ is the vector of all wide model weights, and $\mathbf{w}_{deep}$ are the weights applied on the final activations $a^{(l_f)}$.

\section{System Implementation}
\begin{figure}[t!]
	\centering
	\includegraphics[width=\columnwidth, trim = 0 20 0 20]{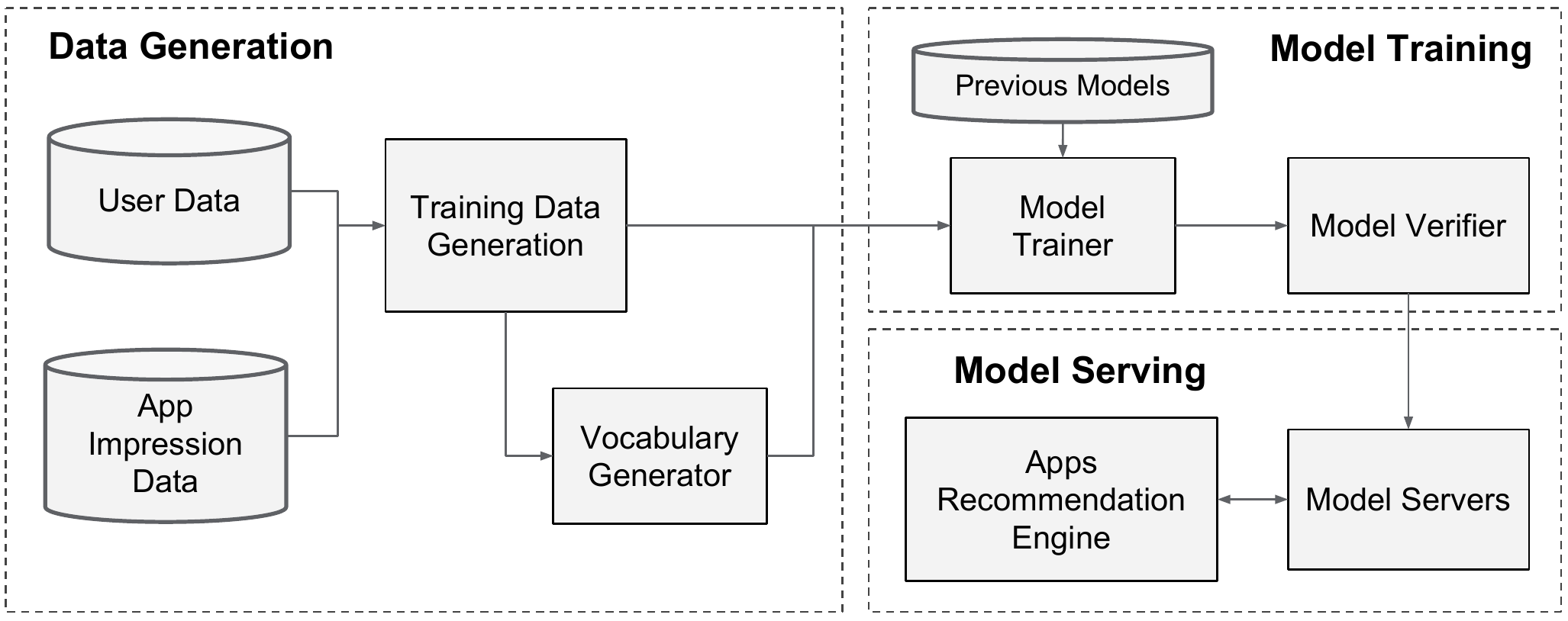}
	\caption{Apps recommendation pipeline overview.}
	\label{fig:PrexSystemImplementation}
\end{figure}

The implementation of the apps recommendation pipeline consists of three stages: data generation, model training, and model serving as shown in Figure \ref{fig:PrexSystemImplementation}.

\subsection{Data Generation}
In this stage, user and app impression data within a period of time are used to generate training data. Each example corresponds to one impression. The label is \textit{app acquisition}: 1 if the impressed app was installed, and 0 otherwise. 

Vocabularies, which are tables mapping categorical feature strings to integer IDs, are also generated in this stage. The system computes the ID space for all the string features that occurred more than a minimum number of times.
Continuous real-valued features are normalized to $[0,1]$ by mapping a feature value $x$ to its cumulative distribution function $P(X \le x)$, divided into $n_q$ quantiles. The normalized value is $\frac{i-1}{n_q-1}$ for values in the $i$-th quantiles. Quantile boundaries are computed during data generation.

\subsection{Model Training}

\begin{figure}[t!]
	\centering
	\includegraphics[width=\columnwidth, trim = 0 20 0 20]{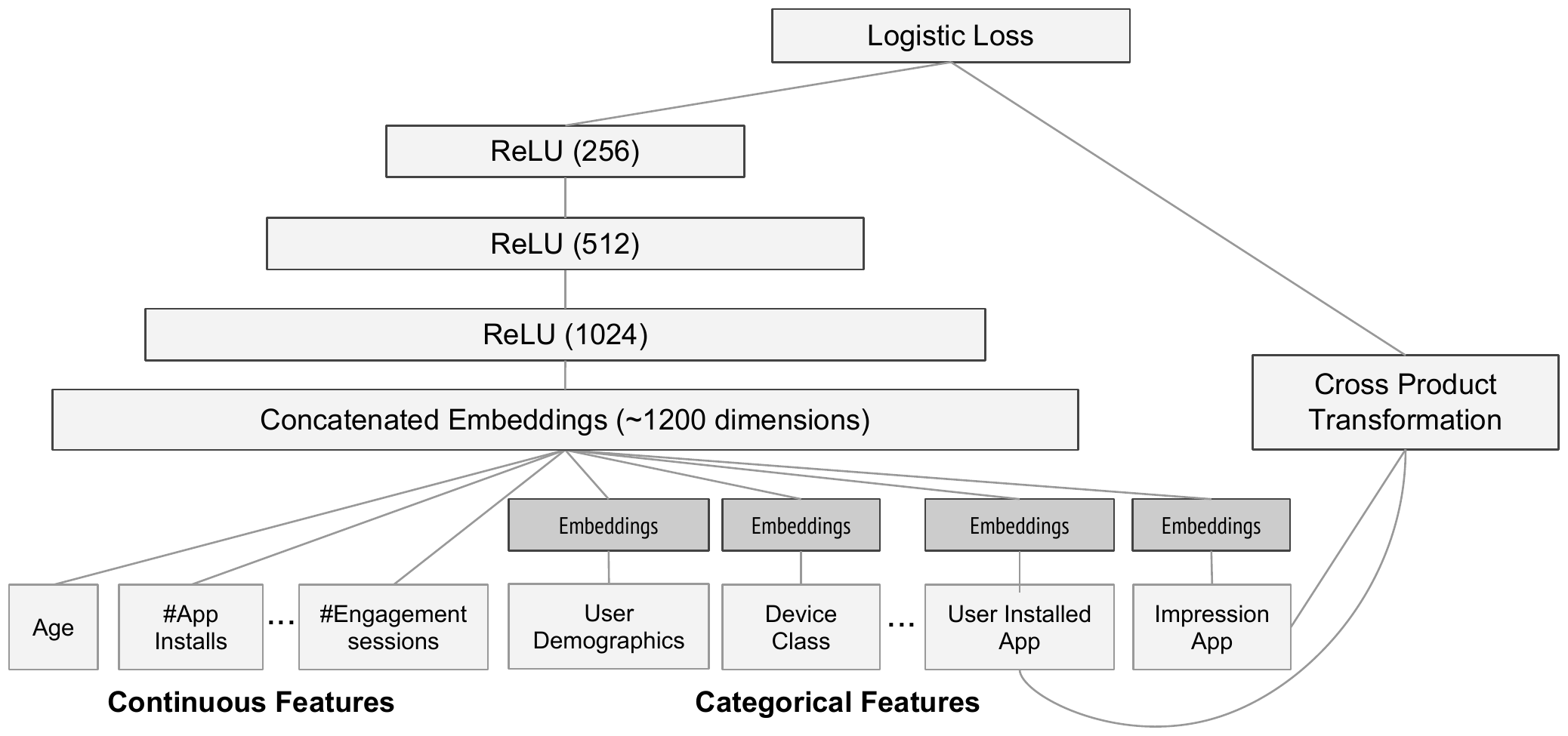}
	\caption{Wide \& Deep model structure for apps recommendation.}
	\label{fig:PrexWideDeepModelStructure}
\end{figure}

The model structure we used in the experiment is shown in Figure \ref{fig:PrexWideDeepModelStructure}. During training, our input layer takes in training data and vocabularies and generate sparse and dense features together with a label. The wide component consists of the cross-product transformation of user installed apps and impression apps. For the deep part of the model, A 32-dimensional embedding vector is learned for each categorical feature. We concatenate all the embeddings together with the dense features, resulting in a dense vector of approximately 1200 dimensions. The concatenated vector is then fed into 3 ReLU layers, and finally the logistic output unit.

The Wide \& Deep models are trained on over 500 billion examples. Every time a new set of training data arrives, the model needs to be re-trained. However, retraining from scratch every time is computationally expensive and delays the time from data arrival to serving an updated model. To tackle this challenge, we implemented a warm-starting system which initializes a new model with the embeddings and the linear model weights from the previous model.

Before loading the models into the model servers, a dry run of the model is done to make sure that it does not cause problems in serving live traffic. We empirically validate the model quality against the previous model as a sanity check.

\subsection{Model Serving}
Once the model is trained and verified, we load it into the model servers. For each request, the servers receive a set of app candidates from the app retrieval system and user features to score each app. Then, the apps are ranked from the highest scores to the lowest, and we show the apps to the users in this order. The scores are calculated by running a forward inference pass over the Wide \& Deep model.

In order to serve each request on the order of 10 ms, we optimized the performance using multithreading parallelism by running smaller batches in parallel, instead of scoring all candidate apps in a single batch inference step.

\section{Experiment Results}
To evaluate the effectiveness of Wide \& Deep learning in a real-world recommender system, we ran live experiments and evaluated the system in a couple of aspects: app acquisitions and serving performance.

\begin{table}
\caption{Offline \& online metrics of different models. Online Acquisition Gain is relative to the control.}
\label{tab:AppAcquisition}
\begin{tabular}{lrr}
\toprule
Model & Offline AUC & Online Acquisition Gain \\ \toprule
Wide (control) & 0.726 & 0\% \\ \midrule
Deep  & 0.722 &  +2.9\%  \\ \midrule
Wide \& Deep  & 0.728 & +3.9\%  \\ \bottomrule
\end{tabular}
\vspace{-13pt}
\end{table}

\subsection{App Acquisitions}
We conducted live online experiments in an A/B testing framework for 3 weeks. For the control group, 1\% of users were randomly selected and presented with recommendations generated by the previous version of ranking model, which is a highly-optimized wide-only logistic regression model with rich cross-product feature transformations. For the experiment group, 1\% of users were presented with recommendations generated by the Wide \& Deep model, trained with the same set of features. As shown in Table \ref{tab:AppAcquisition}, Wide \& Deep model improved the app acquisition rate on the main landing page of the app store by +3.9\% relative to the control group (statistically significant). The results were also compared with another 1\% group using only the deep part of the model with the same features and neural network structure, and the Wide \& Deep mode had +1\% gain on top of the deep-only model (statistically significant). 

Besides online experiments, we also show the Area Under Receiver Operator Characteristic Curve (AUC) on a holdout set offline. While Wide \& Deep has a slightly higher offline AUC, the impact is more significant on online traffic. One possible reason is that the impressions and labels in offline data sets are fixed, whereas the online system can generate new exploratory recommendations by blending generalization with memorization, and learn from new user responses.

\subsection{Serving Performance}
Serving with high throughput and low latency is challenging with the high level of traffic faced by our commercial mobile app store. At peak traffic, our recommender servers score over 10 million apps per second. With single threading, scoring all candidates in a single batch takes 31 ms. We implemented multithreading and split each batch into smaller sizes, which significantly reduced the client-side latency to 14 ms (including serving overhead) as shown in Table \ref{tab:ServingLatency}.

\section{Related Work}
The idea of combining wide linear models with cross-product feature transformations and deep neural networks with dense embeddings is inspired by previous work, such as factorization machines \cite{LibFMTIST12} which add generalization to linear models by factorizing the interactions between two variables as a dot product between two low-dimensional embedding vectors. In this paper, we expanded the model capacity by learning highly nonlinear interactions between embeddings via neural networks instead of dot products.

In language models, joint training of recurrent neural networks (RNNs) and maximum entropy models with \textit{n}-gram features has been proposed to significantly reduce the RNN complexity (e.g., hidden layer sizes) by learning direct weights between inputs and outputs \cite{MaxentRNN11}.
In computer vision, deep residual learning \cite{DeepResidualLearning} has been used to reduce the difficulty of training deeper models and improve accuracy with shortcut connections which skip one or more layers.
Joint training of neural networks with graphical models has also been applied to human pose estimation from images \cite{JointCNNMRF14}. In this work we explored the joint training of feed-forward neural networks and linear models, with direct connections between sparse features and the output unit, for generic recommendation and ranking problems with sparse input data.

In the recommender systems literature, collaborative deep learning has been explored by coupling deep learning for content information and collaborative filtering (CF) for the ratings matrix \cite{CollaborativeDLRecsKDD15}. There has also been previous work on mobile app recommender systems, such as AppJoy which used CF on users' app usage records \cite{AppJoyMobiSys11}. Different from the CF-based or content-based approaches in the previous work, we jointly train Wide \& Deep models on user and impression data for app recommender systems.

\section{Conclusion}
Memorization and generalization are both important for recommender systems. Wide linear models can effectively memorize sparse feature interactions using cross-product feature transformations, while deep neural networks can generalize to previously unseen feature interactions through low-dimensional embeddings. We presented the Wide \& Deep learning framework to combine the strengths of both types of model. We productionized and evaluated the framework on the recommender system of Google Play, a massive-scale commercial app store. Online experiment results showed that the Wide \& Deep model led to significant improvement on app acquisitions over wide-only and deep-only models.

\begin{table}
\caption{Serving latency vs. batch size and threads.}
\label{tab:ServingLatency}
\begin{tabular}{rrr}
\toprule
Batch size & Number of Threads & Serving Latency (ms)\\ \toprule
200 & 1 & 31 \\ \midrule
100 &  2 & 17 \\ \midrule
50 & 4 & 14 \\ \bottomrule
\end{tabular}
\vspace{-13pt}
\end{table}

\bibliographystyle{abbrv}
\bibliography{wide-n-deep-paper-draft}

\end{document}